\documentclass[10pt]{article}
\usepackage[utf8]{inputenc}
\usepackage{amsmath}
\usepackage{amsfonts}
\usepackage{amssymb}
\usepackage{changepage}
\usepackage{bm}
\usepackage[english]{babel}
\usepackage{natbib}
\usepackage{graphicx}
\usepackage{hyperref}
\usepackage[ruled,linesnumbered]{algorithm2e}
\usepackage{mathtools}
\usepackage{float}
\usepackage{threeparttable}
\usepackage{makecell}
\usepackage{diagbox}
\usepackage{subfloat}

\usepackage{color}
\definecolor{dkgreen}{rgb}{0,0.65,0}
\definecolor{gray}{rgb}{0.5,0.5,0.5}
\definecolor{mauve}{rgb}{0.8,0,0.5}
\usepackage{listings}
\setcitestyle{authoryear}

\SetArgSty{textnormal}
\renewcommand{\thefootnote}{\fnsymbol{footnote}}
\graphicspath{ {./} }
\usepackage[verbose=true,letterpaper]{geometry}
\AtBeginDocument{
  \newgeometry{
    textheight=9in,
    textwidth=5.5in,
    top=1in,
    headheight=12pt,
    headsep=25pt,
    footskip=30pt
  }
}
\newcommand{\x}{\bold{x}}
\newcommand{\z}{\bold{z}}
\newcommand{\xpred}{\hat{\x}_\theta(\z)}
\newcommand{\xtilde}{\tilde{\x}}
\newcommand{\boldc}{\bold{c}}
\newcommand{\pz}{p(\z)}
\newcommand{\pxz}{p(\x \vert \z)}
\newcommand{\qsingle}{q(\z \vert \x)}
\newcommand{\posterior}{q(\z_i \vert \z_{<i}, \x)}
\newcommand{\prior}{p(\z_i \vert \z_{<i})}
\newcommand{\brackets}[1]{\left[#1\right]}
\newcommand{\pars}[1]{\left(#1\right)}
\newcommand{\Expect}[2]{\mathbb{E}_{#1}\brackets{#2}}
\newcommand{\KLdiv}[2]{\text{KL}({#1} \Vert {#2})}
\newcommand{\qzx}{q(\z \vert \x)}
\newcommand{\sigmaout}{\sigma_\text{output}}
\newcommand{\bmu}{\boldsymbol{\mu}}
\newcommand{\bsigma}{\boldsymbol{\sigma}}
\newcommand{\diag}[1]{\text{diag}\pars{#1}}
\newcommand{\guided}{\text{guided}}

\newcommand{\conddist}{p(\z_i \vert \z_{<i}, \boldc)}
\newcommand{\unconddist}{p(\z_i \vert \z_{<i})}
\newcommand{\pguided}{p_\text{guided}(\z_i \vert \z_{<i}, \boldc)}
\newcommand{\wmu}{w_\mu}
\newcommand{\wsigma}{w_\sigma}
\newcommand{\sigmac}{\bsigma_\text{c}}
\newcommand{\sigmau}{\bsigma_\text{u}}
\newcommand{\varc}{{\bsigma_\text{c}}^2}
\newcommand{\varu}{{\bsigma_\text{u}}^2}
\newcommand{\muc}{\bmu_\text{c}}
\newcommand{\muu}{\bmu_\text{u}}

\begin{document}

\title{Optimizing Hierarchical Image VAEs for Sample Quality}

\author{
  Eric Luhman\thanks{Equal Contribution} \\
  \texttt{ericluhman2@gmail.com}
  \and
  Troy Luhman\footnotemark[1] \\
  \texttt{troyluhman@gmail.com}
}
\date{\vspace{-5ex}}

\maketitle 

\renewcommand{\thefootnote}{\arabic{footnote}}

\begin{abstract}
While hierarchical variational autoencoders (VAEs) have achieved great density estimation on image modeling tasks, samples from their prior tend to look less convincing than models with similar log-likelihood. We attribute this to learned representations that over-emphasize compressing imperceptible details of the image. To address this, we introduce a KL-reweighting strategy to control the amount of information in each latent group, and employ a Gaussian output layer to reduce sharpness in the learning objective. To trade off image diversity for fidelity, we additionally introduce a classifier-free guidance strategy for hierarchical VAEs. We demonstrate the effectiveness of these techniques in our experiments. Code is available at \url{https://github.com/tcl9876/visual-vae}.

\end{abstract}

\section{Introduction}
\label{section:1}
Deep likelihood based models have achieved impressive capabilities on unsupervised image tasks. Models such as autoregressive models, diffusion models \citep{ddpm}, and variational autoencoders \citep{autoencoding, laddervae} all perform excellently on the log-likelihood metric \citep{sparsetransformer, vdm, nvae}, indicating the promise of each approach. Autoregressive and diffusion models have additionally proved capable of generating high-fidelity images \citep{vqvae, adm}, reaching unprecedented levels of performance in complex text-to-image tasks \citep{dalle, ldm, dalle2, parti}.

While images sampled from autoregressive or diffusion priors usually exhibit a high degree of realism, the same often cannot be said of VAEs despite their good likelihood. This result is especially surprising considering the close similarities between VAEs and diffusion models. Both models optimize a variational inference objective, employ a stack of diagonal Gaussian latent distributions, and exhibit coarse-to-fine generation behavior \citep{ddpm, vdvae}. The primary difference between them is the use of learned posterior distributions in VAEs, compared to the manually specified posteriors in diffusion models. Given the gap in performance, one might wonder if fixed posteriors are indeed better suited for high-fidelity image synthesis.

We begin our paper by offering insights on why existing VAEs are unable to produce high quality samples despite their good likelihood. Specifically, the structure of an image needs only a few bits of information to encode, with the majority being occupied by minuscule details. We argue that hierarchical VAEs are naturally inclined to model these fine details, and can largely ignore global structure since it contributes relatively little to the likelihood.

In this work, we are motivated by the single goal of improving the perceptual quality of samples from the prior of hierarchical VAEs. To this end, we propose two techniques to emphasize modeling global structure. The first technique offers control over the amount of information in each latent group by reweighting terms of the ELBO, which can be used to allocate more latent groups to the first few bits of information. We also replace the discretized mixture of logistics output layer with a Gaussian distribution trained with a continuous KL objective, which greatly reduces the KL used while maintaining near perfect reconstructions. Orthogonal to these techniques, we also introduce a classifier-free guidance strategy for VAEs that trades image diversity for fidelity at sampling time.

We test our method on the CIFAR-10 dataset, showing that our techniques improve the visual quality of generated samples, reducing FID by up to $2\times$ over a controlled baseline. We additionally demonstrate its superior compression capabilities at low rate as further justification of our method despite worse likelihoods. Finally, we verify the effectiveness of classifier-free guidance on class-conditional ImageNet $64^2$.

\section{Background}
\label{section:2}
\subsection{Hierarchical Variational Autoencoders}
\label{section:2.1}

We provide a brief review of hierarchical VAEs in this section; a more thorough introduction can be found in Appendix \ref{appendix:A}. A hierarchical variational autoencoder is a generative model that uses a sequence of latent variables $\z \coloneqq \{\z_1, \ldots, \z_N \}$ to estimate a joint distribution $p(\x, \z) = \pxz p(\z)$, with $\pz \coloneqq \prod_{i=1}^{N} \prior$. In general, the true posterior $p(\z \vert \x)$ is intractable, so an approximate posterior $\qzx \coloneqq \prod_{i=1}^{N} \posterior$ is used instead. The generative model is trained to minimize the following variational inference objective:
\begin{equation}
\label{equation:1}
L \coloneqq \Expect{\qzx}{-\log \pxz} + \underbrace{\KLdiv{q(\z_1 \vert \x)}{p(\z_1)}}_{L_1} + \sum_{i=2}^{N} \underbrace{\Expect{q(\z_{<i} \vert \x)}{\KLdiv{\posterior}{\prior}}}_{L_i}
\end{equation}

Generally, the posterior $\posterior$ is typically a trainable diagonal Gaussian that is learned via the reparamerization trick \citep{autoencoding, rezende}. This differs from autoregressive and diffusion models, which optimize a similar variational bound, only their posterior is untrainable and has fixed dimensionality. 

The objective in Equation \ref{equation:1} can be interpreted as a lossless codelength of the data $\x$, where the $-\log \pxz$ term corresponds to the distortion measured in nats, while the KL terms $L_i=\Expect{q(\z_{<i} \vert \x)}{\KLdiv{\posterior}{\prior}}$ make up the rate \citep{vlae}. Intuitively, the distortion term incentivizes the model to accurately reconstruct the data, while the KL terms encourage it to do so with as little information as possible.   

\subsection{Considerations When Directly Optimizing the ELBO}
\label{section:2.2}

When implementing hierarchical image VAEs with neural networks, a leading strategy is to have latent variables $\z_i$ start at low resolution and increase in resolution \citep{vdvae, nvae}. This choice reflects the inductive bias that image generation should be done in a coarse to fine manner, starting from low-level structure and progressively adding finer details. 

Additionally, since low-level features require much less information to encode than high-level features, we might expect the amount of KL in later latent groups to be significantly higher than in earlier groups. A case otherwise could indicate poor compression of low-level features. This is particularly important when generating visually appealing samples, because small differences between the aggregated posterior and the prior can cause many prior samples to fall outside of the posteriors encountered in training. For example, \cite{d2c} showed how even a single bit of KL between $q(\z_i \vert \z_{<i})$ and $\prior$ can create as much as a 50\% prior hole. For stochastic layers that encode image structure, this prior hole would lead to many structurally incoherent images being sampled. 

Despite these low-quality samples, the log-likelihood would be virtually unchanged, considering most natural images take several thousands bits or more to encode. We hypothesize that during optimization, VAEs are naturally inclined to focus on modeling high-level features that constitute the vast majority of the model's code lengths, weakening their ability to model low-level features. While allocating an equal number of stochastic layers to low, intermediate, and high-level features might be best for sample quality, a model optimized for likelihood would direct most its layers towards encoding high-level features. Such behavior has empirically been observed in diffusion models, where \cite{vdm} found that assigning more stochastic layers towards modeling imperceptible perturbations improved likelihood at the expense of sample quality.

\section{Techniques for Improving Sample Quality}
\label{section:3}
\subsection{Controlling the Amount of Information in Each Layer}
\label{section:3.1}

As discussed in Section \ref{section:2.2}, it might be beneficial if VAEs allocated more stochastic layers for the first few bits of information, which encode important aspects of image structure. However, optimizing Equation \ref{equation:1} with gradient descent will generally lead to most stochastic layers being allocated to high-level features. While many prior works have introduced techniques to prevent layers from encoding zero information, i.e. posterior collapse, \citep{laddervae, vlae, dvaepp}, none offer fine-grained control over the amount of information in each. 

We are interested in learning VAE posteriors that follow a certain ``information schedule" which specifies the desired amount of information in each stochastic layer relative to the total amount. To facilitate a desirable information schedule, we propose to reweight the KL terms of the ELBO based on their value relative to the target KL determined by the information schedule. The weighted objective is of the form 
\begin{equation}
L \coloneqq \Expect{\qzx}{-\log \pxz} + \sum_{i=1}^{N} \lambda(L_i, a, b) L_i
\end{equation}
where we choose $a = \frac{2}{3} l_{target_i} \sum_i L_i$, $b = \frac{4}{3} l_{target_i} \sum_i L_i$, and $l_{target_1}, l_{target_2}, \ldots l_{target_N}$ is a pre-specified sequence of positive constants such that $\sum_i l_{target_i} = 1$. Intuitively, $l_{target_i}$ represents how much KL the $i$-th latent group should contain relative to the total KL; we choose it to be relative because some images inherently require more KL than others, and set a range of $[a,b]$ to give the posterior flexibility. The weighting function is defined as:
\begin{equation}
\lambda(L_i, a, b) = 
	\begin{cases} 
      \max(L_i/a, 0.1) & L_i<a \\
      1 & a\leq L_i\leq b \\
      1+\min((L_i-b)/a), 1) & L_i>b
   \end{cases}
\end{equation}

If the current KL is within the target range, we use the normal weighting $\lambda=1$. As the KL decreases below the lower target, we downweight it to encourage the model to use more information in this latent group; this downweighting factor becomes stronger the farther the KL is from the target. Similarly, if the KL is above the maximum target, we upweight it to discourage use of this group. When implementing this in practice, we apply a stop gradient to the weighting function to make it non-differentiable with respect to the model parameters, and constrain $\lambda$ to be between 0.1 and 2. As for the choice of $l_{target_i}$, we choose it to be an increasing geometric sequence plus a small constant, where $l_{target_N}$ is about 100 times larger than $l_{target_1}$. More specific details can be found in Appendix \ref{appendix:B2}.

\begin{figure}[!htb]
   \begin{minipage}{0.48\textwidth}
     \centering
		\includegraphics[scale=0.37]{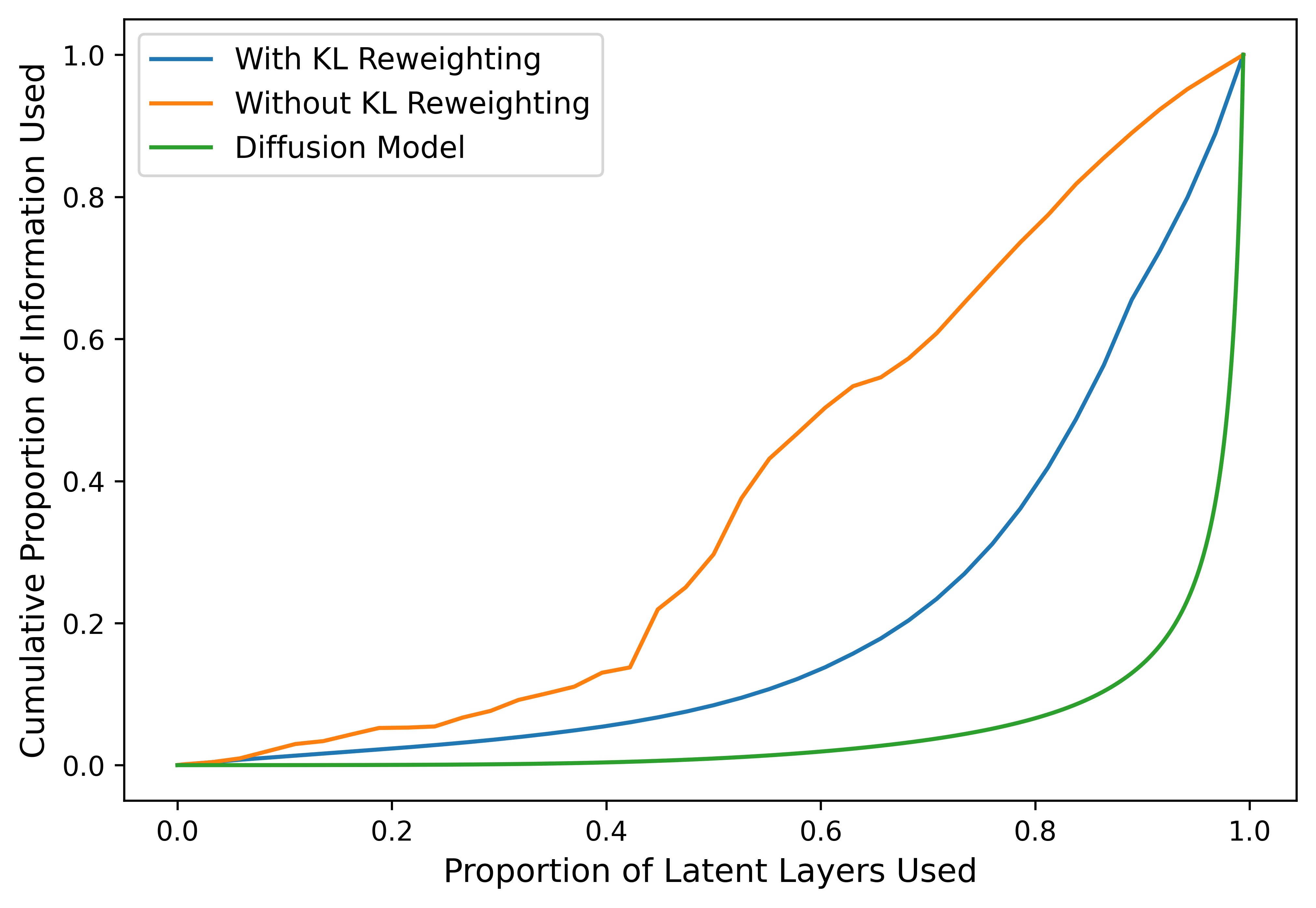}
     \caption{Information schedules for VAEs with and without KL reweighting, compared to the diffusion model from \cite{ddpm} on the CIFAR-10 dataset.}\label{fig:1}
   \end{minipage}\hfill
   \begin{minipage}{0.48\textwidth}
     \centering
		\includegraphics[scale=0.37]{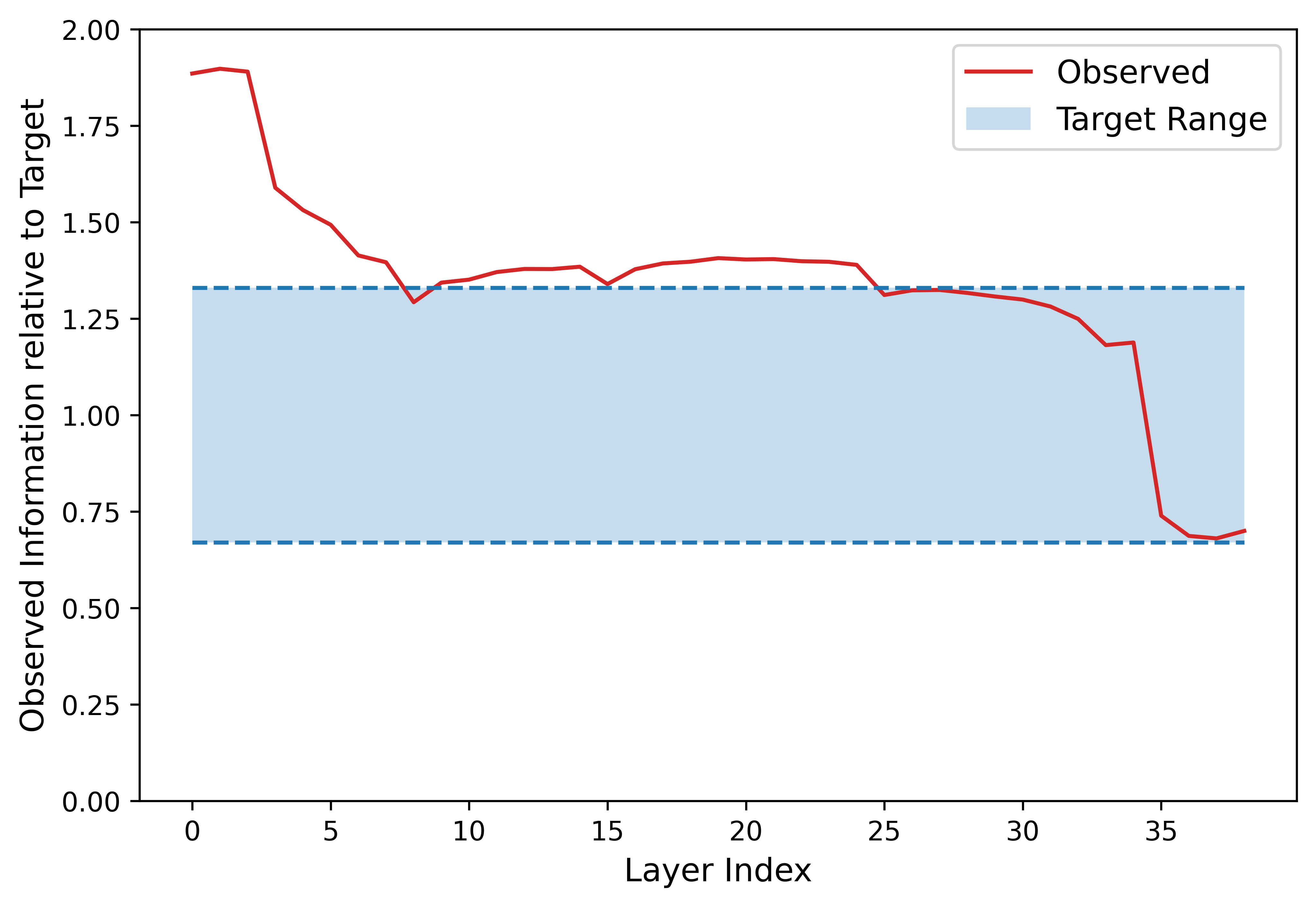}
     \caption{Ratio of the observed KL in each stochastic layer compared to the target range determined by the information schedule. Most fall within the range or slightly above.}\label{fig:2}
   \end{minipage}
\end{figure}

Figure \ref{fig:1} shows the cumulative percentage of information used at each stochastic layer for VAEs with and without our KL-reweighting strategy, and a diffusion model for comparison. Without reweighting the KL terms, most information is added in the middle layers. With the information schedule, it follows a much steeper schedule that adds most the information in the last few stochastic layers. This makes the model allocate less capacity towards modeling high-level features, and more towards global structure. This manner of adding information bears closer similarity to a diffusion model, and we hypothesize such behavior is beneficial to the success of both.

\subsection{Improving Learning Signal with Gaussian Decoders}
\label{section:3.2}

Previous work in hierarchical VAEs have achieved very good image reconstructions, but relatively poor samples from the prior. For instance, NVAE reconstructions on the CIFAR-10 train set achieve a FID \citep{fid} of 2.67, but unconditional samples achieve a FID of 51.71, indicating a large prior hole. We hypothesize that the gap between reconstruction and samples stems from the discrete log-likelihood parameterization of the reconstruction loss. Specifically, the 8-bit log-likelihood term requires performing almost perfect reconstructions to achieve low distortion; a model attaining a reconstruction loss of 1.8 bpd\footnote{\cite{nvae} report a reconstruction loss of 1.8 bits per dim on the CIFAR-10 training set.} must assign a geometric average of 29\% probability to the exact pixel out of 255 possible values. 

We are interested in whether a squared-error reconstruction loss would lead to better learning signal. There might be several reasons for this. Firstly, de-emphasizing the reconstruction loss would in turn cause a decrease in KL and a smaller prior hole. Additionally, a squared-error loss acts in continuous space, which might be more natural for image color values than a discrete log-likelihood loss. 

Optimizing a squared error loss of the form $\gamma ||\x-\xpred||$ is equivalent to minimizing the KL divergence between the distributions $q(\xtilde \vert \x) \coloneqq \mathcal{N} (\xtilde; \x, \sigmaout^2 \textbf{I})$ and $p(\xtilde \vert \z) \coloneqq \mathcal{N} (\xtilde; \xpred, \sigmaout^2 \textbf{I})$ where $\sigmaout = \frac{1}{ \sqrt{2 \gamma}}$. This form of the reconstruction loss more closely resembles the other KL terms in the loss objective. In our experiments, we opt to let the prior learn the variance of $p(\xtilde \vert \z)$ with a diagonal Gaussian distribution $\Sigma_\theta(\z) = \diag{\bsigma_\theta(\z)}$. Our new optimization objective becomes
\begin{equation}
L \coloneqq \Expect{\qzx}{\KLdiv{q(\xtilde \vert \x)}{p(\xtilde \vert \z)}} + \sum_{i=1}^{N} \lambda(L_i, a, b) L_i
\end{equation}

In our experiments, we set $\sigmaout = 0.025$ on data that has been scaled to [-1, 1], or about 3.2 pixels for a [0, 255] scale. We choose this value to encourage sharp reconstructions that appear perceptually the same, while still allowing for significantly more room for error when it comes to predicting the exact pixel. Nevertheless, this choice of $\sigmaout$ upweights the reconstruction loss by nearly three orders of magnitude compared to the $L_2$ objective with $\gamma=1$, leading to reconstructions and samples that are much less blurry. 

While hierarchical VAEs commonly parameterize the $\pxz$ term with a discretized mixture of logistics (DMoL) layer \citep{pixelcnnpp, iafvae}, our neural network outputs means and variances of a continuous Gaussian distribution. As such, we parameterize the $\pxz$ term using a Gaussian CDF function that corresponds to the probability of a sample from $p(\xtilde \vert \z)$ landing in the correct bin, as done in \cite{ddpm}:
\begin{equation}
\label{equation:5}
\begin{split}
& p(\x \vert \z) = \prod_{i=1}^{D} \int_{\delta_{-}(\x_i)}^{\delta_{+}(\x_i)} \mathcal{N}(x; \xpred_i, \bsigma_\theta(\z)_i) \text{d}x
\\
& 
\delta_{-}(x)  =
\begin{cases}
	-\infty & x=-1 \\
    x-\frac{1}{255} & x>-1 \\
\end{cases} \ \ \ \ \ \ \ \delta_{+}(x) = 
\begin{cases}
	\infty & x=1 \\
    x+\frac{1}{255} & x<1 \\
\end{cases} 
\end{split}
\end{equation}
where $D$ is the data dimensionality and subscript $i$ denotes the $i$-th dimension. To generate samples, one could randomly sample from $p(\xtilde \vert \z)$ and display these. However, sampling from this distribution essentially adds random noise to the predicted image, which would hurt visual quality. As such, we output only the predicted mean when displaying samples.

\subsection{Classifier-free guidance in Conditional VAEs}
\label{section:3.3}

Because of their inclusive KL divergence objective, imperfect likelihood-based models will assign high probability to low-density regions of the data distribution; samples from these regions result in low quality images. As such, we are interested in a way to improve fidelity at the expense of diversity. One technique that has recently achieved great success in diffusion models is classifier-free guidance \citep{cfg, glide}. For a conditional model $p(\x \vert \boldc)$, this sampling technique draws samples from $\frac{1}{Z} p(\x \vert \boldc) (\frac{p(\x \vert \boldc)}{p(\x)})^w \propto p(\x \vert \boldc) p(\boldc \vert \x)^w$; which reweights the data distribution according to how likely a sample can be classified as the correct class. This classification uses the model itself to estimate conditional and unconditional probabilities, avoiding the need for an external classification network.

To facilitate guided sampling in VAEs, we first drop the class label in the prior with 10\% probability during training to learn unconditional prior transitions $\unconddist$. During sampling, we keep two separate running hidden states for the conditional and unconditional generation paths, which output latent distributions $\conddist = \mathcal{N} \pars{\z_i; \muc, \diag{\varc}}$ and $\unconddist = \mathcal{N} \pars{\z_i; \muu, \diag{\varu}}$ respectively. The unconditional generation path uses a dummy label as its ``conditioning". From the computed conditional and unconditional distribution parameters, we define a guided probability distribution $\pguided$, and draw a single sample $\z_i \sim \pguided$ that is fed back into both conditional and unconditional hidden states via a linear projection. This process is repeated for each stochastic layer until generation is completed. We provide a PyTorch implementation template in Appendix \ref{appendix:C2}.

The most straightforward approach is to set $\pguided \propto \conddist  \pars{\frac{\conddist}{\unconddist}}^w$, but under certain conditions for $\sigmau$ and $\sigmac$ this particular choice of $\pguided$ fails to be a valid probability distribution (See Appendix \ref{appendix:C1} for details). Instead, we shift and scale the probability density by extrapolating the means and variances in the following manner:
\begin{equation}
\label{equation:6}
\pguided \coloneqq \mathcal{N} \pars{\z_i; \muc + \wmu\left(\muc-\muu\right), \diag{ \varc\left({\frac{\sigmac}{\sigmau}} \right)^{2\wsigma}} }
\end{equation} 

where $\wmu$ and $\wsigma$ are scalars that control how strongly we guide the means and variances of the guided distribution. Under this formulation, we are guaranteed to sample from a valid probability distribution for all positive $\sigmau$ and $\sigmac$. Notably, the values of $\wmu$ and $\wsigma$ need not be equivalent, and can be chosen independently from one another.

\section{Experimental Results}
\label{section:4}

In this section, we evaluate how our techniques affect both sample quality and likelihood. We use the FID metric to assess perceptual quality \citep{fid}, and Precision and Recall metrics to assess fidelity and diversity \citep{precrecall}. 

\subsection{Ablation Studies}
\label{section:4.1}

We ablate the effect of our information scheduling and Gaussian output layers on the unconditional CIFAR-10 dataset \citep{cifar}. Table 1 shows the FID, precision, recall, and likelihood for a baseline hierarchical VAE, one with a Gaussian decoder, one with an information schedule, and one that uses both. Our baseline model is a very deep VAE with similar settings to \cite{vdvae}, that uses regular residual blocks instead of bottleneck blocks, among other changes listed in Appendix \ref{appendix:B1}. This model achieves a FID of 40.65 and a test set log-likelihood of 3.10 bpd. For comparison, the very deep VAEs from \cite{vdvae} obtain an average FID of 37.51 and log-likelihood of 2.87 bpd\footnote{FID calculation was obtained by averaging over the 4 models provided in their codebase.}.

\begin{table}[h!]
  \begin{center}
    \caption{Metrics on CIFAR-10. For models using Gaussian output layers, the value in parenthesis indicates sampling the output layer at full temperature (adding Gaussian noise). All stochastic layers are sampled with temperature 1.} 
    \label{tab:hparams}
    \begin{tabular}{l c c c c} 
      \\ 
      Technique & FID $\downarrow$  & Precision $\uparrow$ & Recall $\uparrow$ & Test NLL (bpd) $\downarrow$ \\
      \hline \\
      Both Techniques & \textbf{20.82} (44.12) & \textbf{0.64} (0.46) & \textbf{0.47} (0.35) & 4.39  \\
      Gaussian Output Layer & 27.49 (51.26) & 0.63 (0.48) & 0.45 (0.34) & 4.38 \\
      Information Schedule & 31.21 & 0.63 & 0.41 & 3.10 \\
      Neither Technique & 40.66 & 0.64 & 0.35 & \textbf{3.10} \\
      
    \end{tabular}
  \end{center}
\end{table}

When applying both proposed techniques, the FID decreases to 20.82, improving over the control by a factor of two. Since model capacity and training time are fixed across experiments, this improvement comes solely from optimizations conducive to sample quality. We also validate the use of each individual technique, where both provide a substantial boosts in FID scores compared to the baseline. Nevertheless, combining them together leads to the greatest improvement in sample quality, indicating the techniques are complementary. This makes sense because each technique acts in a different way; one affects the way information is added, while the other changes the parameterization of the reconstruction term.

While our proposed techniques significantly improve sample quality metrics, the same cannot be said about the likelihood. In particular, replacing the DMoL output layer with a Gaussian one drastically worsens likelihood. This is caused by a very high distortion term, which in this case makes up about 3.75 of the roughly 4.39 bpd. In turn, the KL used by the model is far less, totaling compared 0.64 bpd compared to the 1.67 used by the baseline model, while attaining similarly accurate reconstructions to human perception (see Appendix \ref{appendix:D}). This lower KL decreases the prior hole, which improves sample quality. 

\subsection{Rate-distortion Comparisons}
\label{section:4.2}
As discussed earlier, using a Gaussian output layer improves quality metrics like FID but greatly worsens test-set likelihood. Thus, one might wonder if the model is learning better representations,
as opposed to simply memorizing training images or creating adversarial examples that the Inception network used in calculating metrics judges favorably \citep{barratt2018note}. To demonstrate otherwise, we analyze the rate-distortion curves of each of our models from Section \ref{section:4.1}, showing that reducing distortion at low rate is key to improving sample quality. 

As discussed in Section \ref{section:2.1}, the model's log-likelihood is analogous to its lossless compression ratio that combines rate and distortion terms. Notably, the log-likelihood metric only measures the compression ratio at the model's maximum rate, when all latent variables are sampled from the posterior. However, a good hierarchical model should also achieve low distortion in the low rate regime, indicating it has learned efficient representations where even a few bits of information can encode global features needed for reconstruction.

\begin{figure}
\begin{center}
\includegraphics[scale=0.7]{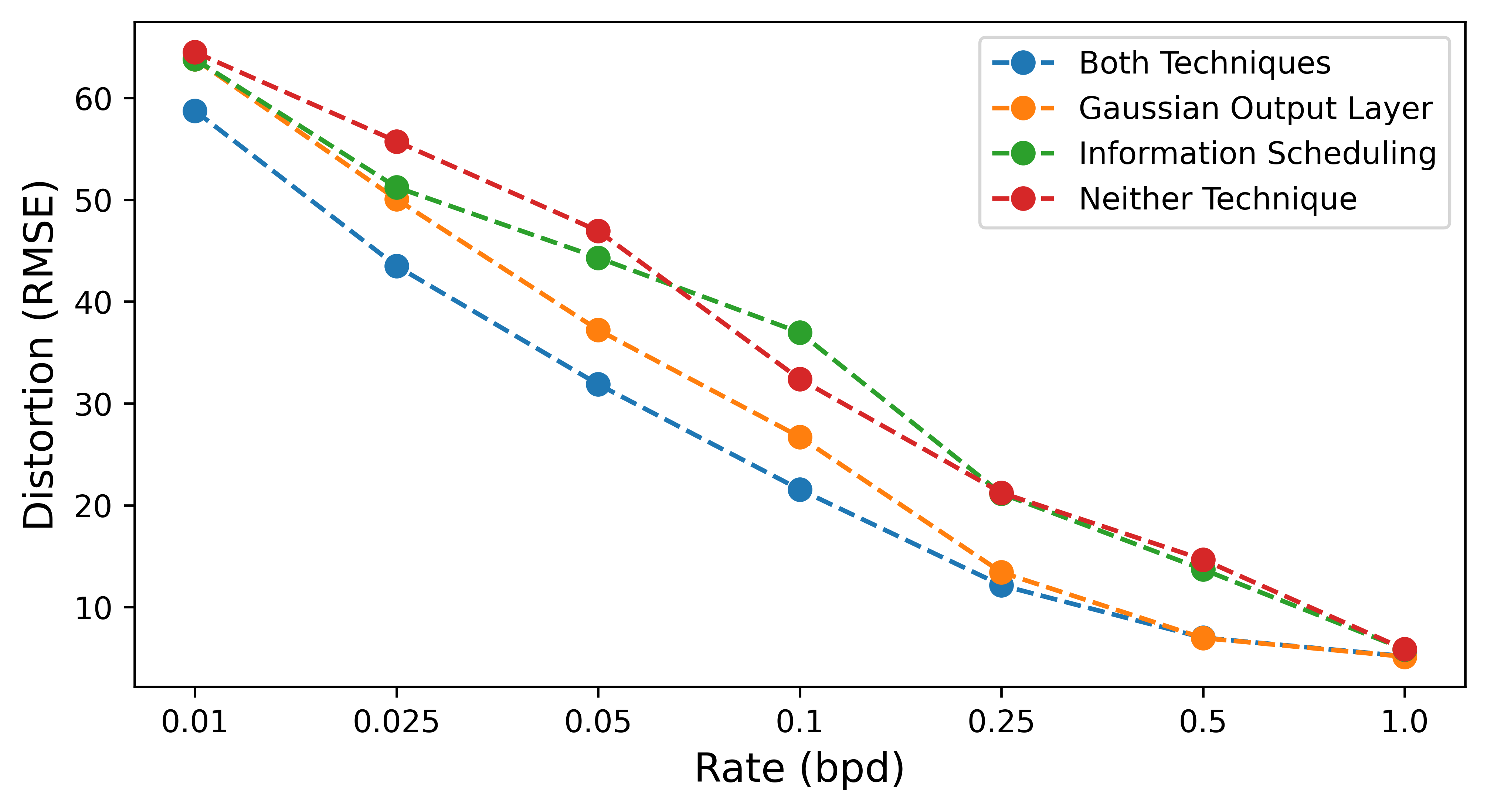}
\caption{Distortion at various rates on the CIFAR-10 test set (values averaged over 10 separate runs to reduce variance caused by stochastic sampling). Our proposed techniques reduce the KL needed to achieve the same reconstruction quality, shrinking the prior hole.}\label{fig:3}
\end{center}
\end{figure}

Figure \ref{fig:3} shows the distortion at various rates for each ablation in Section \ref{section:4.1}. To facilitate these ``partial reconstructions'', we sample from the latent variables from the posterior until we've reached the allotted information, then sample the rest from the prior. Distortion is calculated by finding the root mean squared error (RMSE) between these samples and the original image.

The figure shows that using Gaussian output layers greatly reduces the distortion when reconstructing at a rate of 0.025 to 0.1 bits per dim (compression ratios of 0.3-1.2\% for data with 8 bits per dim). Attaining good reconstructions at these extreme compression ratios is more correlated with good sample quality than likelihood, since the model must learn effective features to compress the image's structural components. 

\subsection{Evaluating classifier-free guidance}
\label{section:4.3}
In this section, we consider a VAE trained on class-conditional ImageNet $64^2$. The VAE is comprised of a $32^2$ generative model and a super-resolution VAE that upsamples from $32^2$ images to $64^2$ images. We chose this parameterization to reduce memory constraints, which can be prohibitively large for very deep models. When conditioning the super-resolution model, add Gaussian noise with $\sigma_\text{aug}^2 > \sigmaout^2$ to the low-resolution input to reduce compounding error resulting from the previous model; this acts similarly to the Gaussian noise conditioning augmentation in \cite{cdm}.


\begin{table}[h!]
  \begin{center}
    \caption{Effect of mean and variance guidance on ImageNet $64^2$ FID/Precision/Recall. Guidance is only applied to the base $32^2$ generator and not in the super-resolution model.} 
    \label{table:2}
    
\begin{tabular}{|l|*{5}{c|}}\hline
\backslashbox{$\wsigma$}{$\wmu$}
&\makebox[3em]{0.0}&\makebox[3em]{0.5}&\makebox[3em]{1.0}&\makebox[3em]{1.5}&\makebox[3em]{2.0}\\\hline
0.0 & 37.0/.49/\textbf{.50} & 30.7/.54/.48 & 26.5/.55/.48 & 24.3/.57/.47 & 23.8/.57/.47\\\hline
0.5 & 35.6/.51/.49 & 29.1/.55/.48 & 24.9/.58/.47 & 22.7/.59/.46 & 22.3/.58/.45 \\\hline
1.0 & 34.5/.53/.48 & 27.8/.57/.47 & 24.0/.59/.47 & 21.5/.60/.44 & 21.1/.61/.44 \\\hline
2.0 & 32.7/.55/.46 & 25.9/.60/.45 & 21.8/.62/.44 & 19.8/.63/.44 & 19.5/.63/.43 \\\hline
3.0 & 31.4/.57/.46 & 24.6/.62/.45 & 20.5/.64/.43 & 18.7/.65/.43 & 18.6/.64/.41 \\\hline
4.0 & 30.5/.58/.45 & 23.6/.63/.43 & 19.7/.65/.42 & 18.0/.66/.41 & 17.9/.65/.39 \\\hline
5.0 & 29.8/.59/.44 & 23.0/.64/.42 & 19.1/.66/.41 & 17.6/\textbf{.67}/.40 & \textbf{17.5}/.66/.39 \\\hline
\end{tabular}

  \end{center}
\end{table}

\begin{figure}[!htb]
   \begin{minipage}{0.5\textwidth}
     \centering
		\includegraphics[scale=0.48]{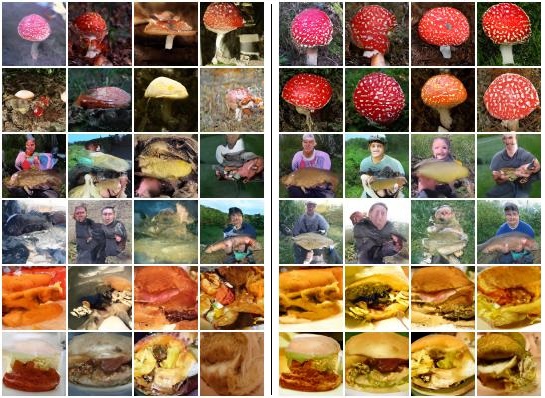}
     \caption{Unguided (left) and guided (right) samples for the agaric, tinca, and cheeseburger classes}\label{fig:4}
   \end{minipage}\hfill
   \begin{minipage}{0.45\textwidth}
     \centering
		\includegraphics[scale=0.37]{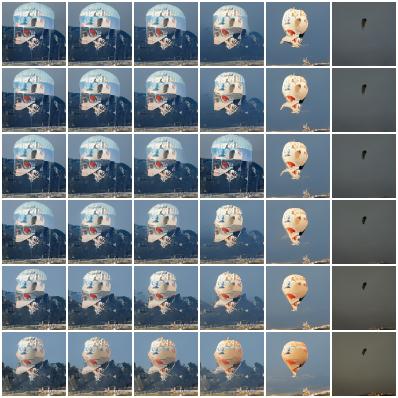}
     \caption{The effect of increasing mean and variance guidance for the balloon class. Columns show $\wmu = \{0, 0.5, 1, 2, 4\}$ and rows show $\wsigma = \{0, 1, 2, 4, 8\}$}\label{fig:5}
   \end{minipage}
\end{figure}

Table \ref{table:2} demonstrates the effect of varying $\wmu$ and $\wsigma$ for our ImageNet model. The effect of guidance is major and decreases FID over the base class-conditional sampler by over 50\%. We find that $\wsigma$ is amenable to higher guidance weights than $\wmu$, as $\wmu$ stops showing improvement after a weight of 1.5, while $\wsigma$ offers improvements even up to a weights of 5. Figure \ref{fig:4} shows the effect of guided sampling; it leads to cleaner samples that more closely resemble their corresponding label. But in doing so, it causes samples to fall unto a few high-density regions, reducing sample diversity. Figure \ref{fig:5} shows the effect of mean and variance guidance independently. Both mean and variance guidance improve the structural cohesion of the image's subject, while simultaneously reducing sources of variation in the background. Additionally, high mean guidance may sometimes induce artifacts, as seen in the last column.

\section{Related Work}
Our work builds off the existing work in hierarchical VAEs \citep{laddervae, iafvae, biva, vdvae}, which serves as the basis for our method. While these works focus on improving the likelihood assigned by the model, our work proposes techniques to improve sample quality at the expense of likelihood. Our information scheduling is enforce via a reweighted ELBO objective that bears similarity to the KL-balanced objective from \cite{dvaepp}. Their reweighting objective prevents posterior collapse by encouraging each stochastic layer to use the same amount of information, and is only used in the beginning of training. Meanwhile, ours allows designating any amount of information to a given latent group with the explicit intent of creating better hierarchical representations.

Our Gaussian output layer is closely related to the $L_2$ error, which is commonly used among practitioners as a reconstruction loss for single-group VAEs. VAEs trained with this objective tend to produce blurry samples, but we avoid this by using a small value of $\sigmaout$. \cite{dai2019diagnosing} provides theoretical arguments to show how a small $\sigmaout$ is beneficial for the log-likelihood. However, to our knowledge this work is the first to explicitly demonstrate the superior performance of a Gaussian output layer over the DMoL layer in terms of sample quality. \cite{distsmooth} is similarly motivated to our work, training an autoregressive model on a data distribution perturbed by Gaussian noise. Their approach requires training a second stage to recover the unperturbed distribution, while ours achieves good performance with only one network. 

Finally, our classifier-free guidance technique is based on a similar approach used in diffusion models \citep{cfg}. However, guided sampling is not as straightforward in VAEs due to their non-Markovian nature, where information about the class label may inadvertantly leak into the hidden state. We emphasize that both our parameterization of the guided distribution in equation \ref{equation:6} and the VAE-specific code implementation for performing guided sampling are novel contributions.

\section{Limitations and Conclusion}

In this work, we focus on optimizing hierarchical VAEs for visual quality, and find that they can successfully generate high-quality images at a level not yet demonstrated by pure VAEs. To do so, we control the amount of information added in each layer, which achieves better compression at low-rate regimes, and use a Gaussian output layer to avoid modeling visually imperceptible details and reduce the KL term. Finally, we demonstrate how guided sampling, an important scaling technique for multimodal datasets, can be applied to VAEs.

Perhaps our method's most significant limitation is the poor likelihood that comes with Gaussian output layers; this trade-off between perceptual quality and likelihood naturally arises when optimizing for the quality of prior samples instead of reconstructions. It is likely that this poor likelihood can be mitigated by using a second model instead of the Gaussian decoder from Equation \ref{equation:5}. Secondly, our techniques introduce more hyperparameters to choose from, such as $l_{target_{1:N}}$ and $\sigmaout$, and hyperparameters that worked for our experiments may not hold for other datasets. On the other hand, an extensive sweep over these hyperparameters can result in even better performance. Lastly, while our contributions make progress in VAE samples' perceptual quality, they have yet to achieve the photorealism of GANs and diffusion models.

In a broader context, we overcome some of the challenges of modeling high-dimensional data with trainable posteriors, demonstrating that the flexible, non-markovian posteriors can learn expressive yet efficient latent variable decompositions. These efficient latent orderings could prove useful in fields like representation learning, and might also result in better model interpretability and more controllable generation.

\section{Acknowledgements}
The authors thank the TPU Research Cloud (TRC) Program for providing Cloud TPUs used in our experiments.

\bibliography{references}

\pagebreak

\appendix
\section{Variational Autoencoders}
\label{appendix:A}

\subsection{Single-group VAEs}
We provide a more thorough introduction to variational autoencoders and their hierarchical variants in this appendix. A variational autoencoder is a generative model that uses latent variables $\z$ to estimate a joint distribution $p(\x, \z) = \pxz p(\z)$. While the probability density can be obtained using Bayes' rule, the true posterior $p(\z \vert \x)$ is usually intractable. However, we can obtain a lower bound on the log-probability $p(\x)$ by introducing an approximate posterior distribution $q(\z \vert \x)$ that is learned jointly with the generative model. 
\begin{align}
\log p(\x) & = \log \pxz + \log p(\z) - \log p(\z \vert \x) \\
& = \log \pxz + \log p(\z) - \log \qsingle - \log p(\z \vert \x) + \log \qsingle 
\end{align}

Taking an expectation over the approximate posterior, we get:
\begin{equation}
\begin{split}
\Expect{\qsingle}{\log p(\x)} = \Expect{\qsingle}{\log \pxz} & + \Expect{\qsingle}{\log p(\z) - \log \qsingle}  \\ & - \Expect{\qsingle}{\log p(\z \vert \x) - \log \qsingle} 
\end{split}
\end{equation}
\begin{align}
\log p(\x) & = \Expect{\qsingle}{\log \pxz} - \KLdiv{\qsingle}{p(\z)} + \KLdiv{\qsingle}{p(\z \vert \x)} \\
\log p(\x) & \geq \Expect{\qsingle}{\log \pxz} - \KLdiv{\qsingle}{p(\z)}
\end{align}
where the last equation holds because the KL divergence is always non-negative. Typically, $\qsingle$ and $p(\z)$ are diagonal Gaussian distributions, which is very useful when using gradient-based optimization techniques. By sampling $\boldsymbol{\epsilon} \sim \mathcal{N}(\textbf{0}, \bold{I})$ and setting $\z = \bmu_\theta(\x) + \boldsymbol{\epsilon} \odot \boldsymbol{\sigma}_\theta(\x)$, we can differentiate the loss terms with respect to $\bmu_\theta$ and $\boldsymbol{\sigma}_\theta$ \citep{autoencoding, rezende}. 

\subsection{Hierarchical VAEs}
Compared to other likelihood based models like energy-based models and normalizing flows \citep{ebm, normflow}, VAEs are advantageous because of they do not require dealing with a partition function or Jacobian determinant. However, fully factorized Gaussian VAEs often struggle to model more complex probability distributions. To gain better expressiveness, Hierarchical VAEs factorize the prior and posterior over groups of latent variables $\z \coloneqq \{\z_1, \ldots, \z_N \}$, with the prior defined as $\pz \coloneqq \prod_{i=1}^{N} \prior$, and the posterior as $\qzx \coloneqq \prod_{i=1}^{N} \posterior$.

Creating a conditional dependence among latent variables leads to a hierarchical generation progression that generates latents at the top of the hierarchy first, and latents later in the hierarchy from those. For image data, earlier latents would naturally encode things like color and shape, while later ones would encode fine details and textures. It would also make sense if the early latents determining structure had lower resolution than the later latents determining details. 

\begin{figure}
\begin{center}
\includegraphics[scale=0.19]{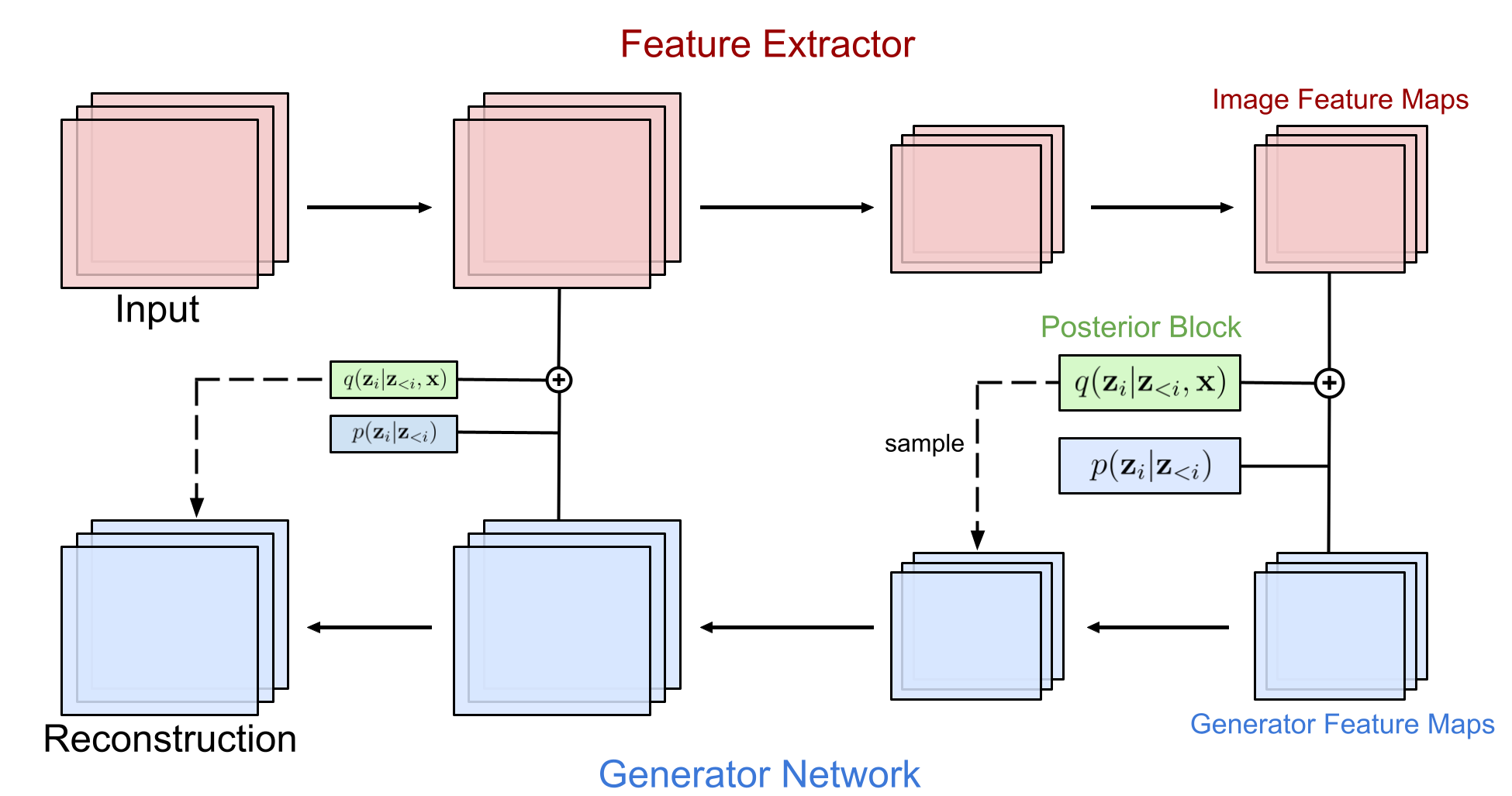}
\caption{Diagram of a hierarchical Image VAE. Solid arrows indicate convolutional connections, while dotted lines indicate connections through stochastic layers.}\label{fig:6}
\end{center}
\end{figure}

A hierarchical VAE implementation consists of three parts: a generator models the prior and performs reconstructions, a series of posterior blocks that take in the current generator hidden state and features of the image being reconstructed, and a feature extractor to provide information about that image to the posterior. The generator progressively adds features from new latent variables, while also processing these features with residual blocks. The generator layers go from low to high resolution, and thus the posterior blocks do as well. However, the image itself is of high resolution, so to get low-resolution inputs we use a feature extractor network to progressively extract lower and lower resolution features. These are then concatenated channel-wise to the generator hidden state. A diagram of this process is illustrated in Figure \ref{fig:6}.

\pagebreak

\section{Experimental Details}

\subsection{Architecture}
\label{appendix:B1}
Our training setup is a variant of the VDVAE architecture, with several significant changes.
\begin{itemize}
\item{To achieve better memory efficiency, we replace the VDVAE bottleneck blocks with BigGAN residual blocks. We also use gradient rematerialization via flax.linen.remat, recomputing activations in residual blocks each time.}
\item{We introduce single-headed attention in the encoder and the generator to better model global relationships.}
\item{Similar to \cite{nvae}, we use spectral regularization \cite{spectralreg} to help stabilize training. However, unlike \cite{nvae}, we only apply it at the layers that produce the mean and variance of the prior and posterior. We also use the gradient skipping heuristic from \cite{vdvae}, but encountered very few skips during training.}
\item{Similar to \cite{efficientvdvae}, we use gradient smoothing, except we parameterize the variance, not the standard deviation, with a softplus function. We use the Adamax optimizer with $\beta_1 = 0.9$ and $\beta_2 = 0.999$, which we found to be more stable \cite{adam}. All models use 50 steps of warmup, and we found higher values of warmup to be unstable.}
\item{For class conditional models, we use separate embeddings for each residual block, which are then linearly projected to a higher channel value. We chose this approach for ease of implementation. Additionally, the encoder and posterior blocks are always class-conditional, even when the label is dropped in the generator for classifier-free guidance.} 
\end{itemize}

\begin{table}[h!]
  \begin{center}
    \caption{Configuration of our cascaded imagenet model.} 
    \label{tab:hparams}
    \begin{tabular}{l c c c c} 
      \\ 
      Dataset & CIFAR-10 & ImageNet $32^2$ & ImageNet $32^2 \rightarrow 64^2$ \\
      \hline \\
      Parameters & 45.4M & 198.5M & 147.5M \\
      Base channel size & 128 & 128 & 128\\
      Channel multiplier & 1,1,1,1,1 & 1,2,2,2,2 & 1,2 \\
      Attention resolutions & None & 16, 32 & 32 \\
      Dropout & 0.0 & 0.0 & 0.0 \\
      Batch size & 128 & 256 & 192 \\
      Max Learning Rate & $6 \times 10^{-4}$ & $4 \times 10^{-4}$ & $3 \times 10^{-4}$ \\
      Min Learning Rate & $3 \times 10^{-4}$ & $2 \times 10^{-4}$ & $3 \times 10^{-4}$ \\
      Training iterations & 350000 & 1052500 & 337500 \\
      $\sigmaout$ & 0.025 & 0.025 & 0.025 \\
      $\sigma_\text{aug}$ & N/A & N/A & 0.1 \\
      SR Penalty & 0.05 & 0.1 & 0.25 \\
      Skip Threshold & 300 & 300 & 200 \\
      EMA Decay & 0.9997 & 0.9999 & 0.9998
      
    \end{tabular}
  \end{center}
\end{table}

\pagebreak

\subsection{Information Schedule and Other Details}
\label{appendix:B2}

The search space for the choice of $l_{target_{1:N}}$ is quite large as it involves choosing an entire sequence of hyperparameters. To simplify matters, we set $l_{target_{1:N}}$ to be $\frac{1}{\Sigma_i l_{target_i}} \exp \pars{\text{linspace} (1, A)}+ B$ as an exponentially increasing sequence plus a constant (however many other options are possible). Intuitively, $A/B$ controls how much more KL is in the last group compared to the first, increasing both $A$ and $B$ relative to $1$ can be used to create a sharper increase at the end of the curve, and the $\frac{1}{\Sigma_i l_{target_i}}$ ensures the values add up to 1. We set $A=1000$ and $B=10$ for our CIFAR-10 and ImageNet $32^2$ models.

In our ImageNet $32^2 \rightarrow 64^2$ super-resolution model, we choose a different approach that sets the target amount of information proportional to the resolution of the latent group, where $64^2$ latent groups are allocated twice as much information as $32^2$ ones. Since a lot of information is already included in the low-resolution image, there's no need to have a steeply increasing information schedule. In general, a suitable choice for $l_{target_{1:N}}$ will reflect inductive biases about how information should be added when generating data, which may be different for depending on the type of data or its dimensionality.  

Models were trained on a mix of TPUv2-8s and TPUv3-8s. When training ImageNet models on TPUv2s, we cast optimizer states to bfloat16, drop calculation of exponentially-moving averaged parameters, and  perform gradient accumulation for our super-resolution model with batch size 64 and 3 accumulations per update. Calculation of EMA parameters is done for the last 50-100 thousand iterations where we train on TPUv3s. Random horizontal flips were used in the CIFAR-10 and ImageNet super-resolution models. FID, precision, and recall metrics are evaluated with code from the guided diffusion repository at \url{https://github.com/openai/guided-diffusion}.

\pagebreak

\section{Classifier-free Guidance}
\label{appendix:C1}
\subsection{Discussion}
As discussed in Section \ref{section:3.3}, the most straightforward way to perform guided sampling would be to set the guided distribution proportional to $\conddist  \pars{\frac{\conddist}{\unconddist}}^w$. Notably, if $\conddist$ and $\unconddist$ are diagonal Gaussian distributions, then $\pguided = \frac{1}{Z} \conddist  \pars{\frac{\conddist}{\unconddist}}^w$ may under certain conditions also be a diagonal Gaussian for some finite normalizing constant $Z$. We will show the derivation for the one-dimensional case where $\mu_c$, $\mu_u$ and $\sigma_c^2$, $\sigma_u^2$ are the means and variances of the conditional and unconditional distributions, and under what conditions $\pguided$ forms a valid probability distribution.

\begin{equation}
p_\guided(z_i \vert z_{<i}, c) = \frac{1}{Z} p(z_i \vert z_{<i}, c) \pars{\frac{p(z_i \vert z_{<i}, c)}{p(z_i \vert z_{<i})}}^w 
\end{equation}

\begin{equation}
 = \frac{1}{Z} \left(\frac{1}{\sqrt{2\pi \sigma_c^2}}\exp\left(-\frac{\left(z_i-\mu_c\right)^{2}}{2\sigma_c^2}\right)\right)^{\left(1+w\right)} {\left(\frac{1}{\sqrt{2\pi \sigma_u^2}}\exp\left(-\frac{\left(z_i-\mu_u\right)^{2}}{2\sigma_u^2}\right)\right)^{-w}}
\end{equation}

\begin{equation}
 = \frac{\sigma_u^w}{Z\sqrt{2\pi}\sigma_c^{(1+w)}}  \exp\left(-\frac{1}{2} \pars{(1+w)\frac{\left(z_i-\mu_c\right)^{2}}{\sigma_c^2}\ -w\frac{\left(z_i-\mu_u\right)^{2}}{\sigma_u^2}}\right) 
\end{equation}

If $p_\guided$ is a normal distribution the term $(1+w)\frac{\left(z_i-\mu_c\right)^{2}}{\sigma_c^2} -w\frac{\left(z_i-\mu_u\right)^{2}}{\sigma_u^2}$ must have the same coefficients in front the $z_i^2$ and $z_i$ terms as $(z_i - \mu_\guided)^2/\sigma_\guided^2$. (These expressions will differ by an additive constant that will affect the value of the normalizing constant $Z$ when taken out of the exp function). Expanding out and rearranging, this expression becomes:

\begin{equation}
z_i^2 \pars{\frac{1+w}{\sigma_c^2} - \frac{w}{\sigma_u^2}} - 2z_i \pars{\mu_c \frac{1+w}{\sigma_c^2} - \mu_u \frac{w}{\sigma_u^2}} + \pars{\mu_c^2 \frac{1+w}{\sigma_c^2} - \mu_u^2 \frac{w}{\sigma_u^2} }
\end{equation}
If we set $\sigma_\guided^2 = \pars{\frac{1+w}{\sigma_c^2} - \frac{w}{\sigma_u^2}}^{-1}$ and $\mu_\guided = \sigma_\guided^2 \pars{\mu_c \frac{1+w}{\sigma_c^2} - \mu_u \frac{w}{\sigma_u^2}}$, the coefficients for the $z_i^2$ and $z_i$ terms will equal those for the $(z_i - \mu_\guided)^2/\sigma_\guided^2$. However, when examining the expression for $\sigma_\guided^2$, we see that it is not positive when $\frac{w}{\sigma_u^2} \geq \frac{1+w}{\sigma_c^2}$. In this case, $p_\guided$ will not form a valid probability distribution for any normalizing constant $Z$, with the positive coefficient in front of the $z_i^2$ term causing the function to diverge. When $\frac{w}{\sigma_u^2} < \frac{1+w}{\sigma_c^2}$, $p_\guided$ would form a Gaussian distribution with mean $\mu_\guided$ and variance $\sigma_\guided^2$. 

Although the extrapolation guidance in Equation \ref{equation:6} ensures the guided distribution is valid for all positive values of $\sigma_c^2$ and $\sigma_u^2$, we did notice saturation artifacts with high guidance weights, and in rare cases NaN values caused by numeric overflow. This occurs when the class-conditioned means and variances becomes significantly larger in magnitude than the unconditional ones, causing $\muc + w(\muc - \muu)$ to essentially act as $(w+1)\muc$ which in turn causes exponentially growing magnitudes.

\pagebreak

\subsection{Example Implementation}
\label{appendix:C2}
We provide an example implementation for guided sampling from one VAE stochastic layer. For the full pipeline, we encourage readers to see our codebase at \url{https://github.com/tcl9876/visual-vae}.

\begin{lstlisting}
# Code to perform guided sampling on a HVAE stochastic layer 
class StochasticLayer(nn.Module):
	...
	
	def sample_guided(self, h_c, h_u, label, w_mean, w_var):
		uncond_label = torch.ones_like(label) * NUM_CLASSES
		
		cond_output = self.conv_block1(h_c, label) 
		uncond_output = self.conv_block1(h_u, uncond_label) 
		pmean_c, plogvar_c, residual_c = torch.split(cond_output, [self.z_channels, self.z_channels, self.hidden_size], dim=1)
		pmean_u, plogvar_u, residual_u = torch.split(uncond_output, [self.z_channels, self.z_channels, self.hidden_size], dim=1)
		
		guided_mean = pmean_c + w_mean * (pmean_c - pmean_u)
		guided_logvar = plogvar_c + w_var * (plogvar_c - plogvar_u)
		z = guided_mean + torch.randn(pmean_c.shape) * torch.exp(0.5 * guided_logvar) 
		
		z = self.z_projection(z)		
		h_c = h_c + residual_c + z
		h_u = h_u + residual_u + z
		h_c = h_c + self.conv_block2(h_c, cond_label)
		h_u = h_u + self.conv_block2(h_u, uncond_label)
		return h_c, h_u
    
\end{lstlisting}

\pagebreak

\section{CIFAR-10 Samples}
\label{appendix:D}
\begin{figure}[h!]
\begin{center}
\includegraphics[scale=0.16]{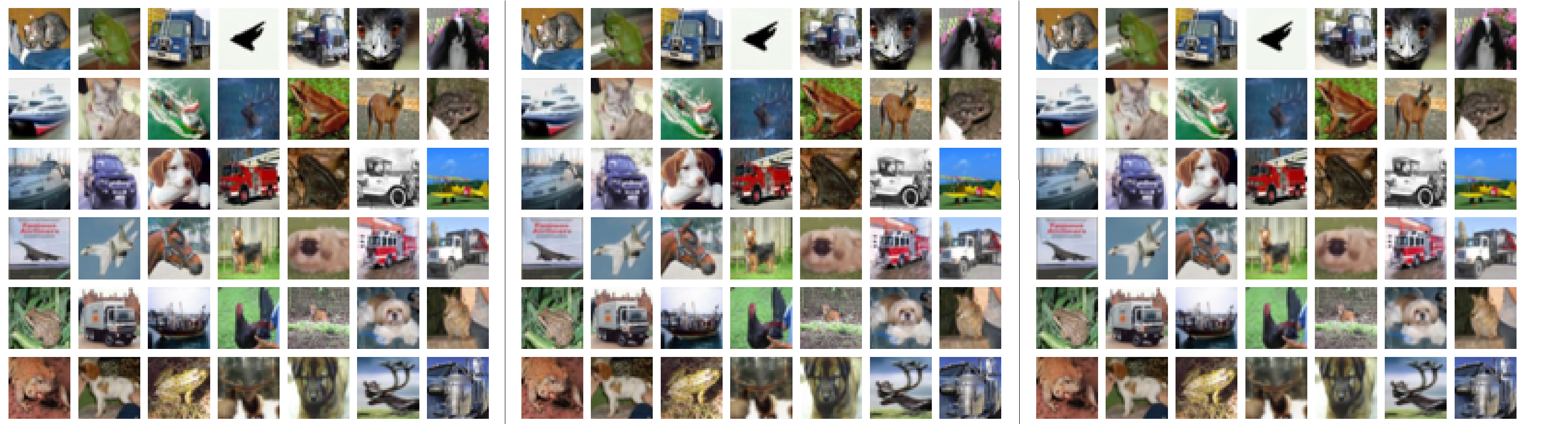}
\vspace{-0.5cm}
\caption{Test set images (left) and reconstructions from a model with no Gaussian output layer (middle, distortion = 1.5 bpd) and with one (right, distortion = 3.75 bpd). The Gaussian output layer is sampled at temperature zero. They appear perceptually the same despite one have much higher reconstruction loss.}\label{fig:8}
\end{center}
\end{figure}

Figure \ref{fig:7} shows prior samples from each ablation from Section \ref{section:4.1}. Samples from our model show significantly better structure than those from the baseline. We also visualize reconstructions in Figure \ref{fig:8}, showing how models with Gaussian output layers achieve sharp reconstructions despite their high reconstruction loss.

\begin{subfigures}
\label{fig:7}
\begin{figure}[h!]
   \begin{minipage}{0.48\textwidth}
     \centering
		\includegraphics[scale=0.17]{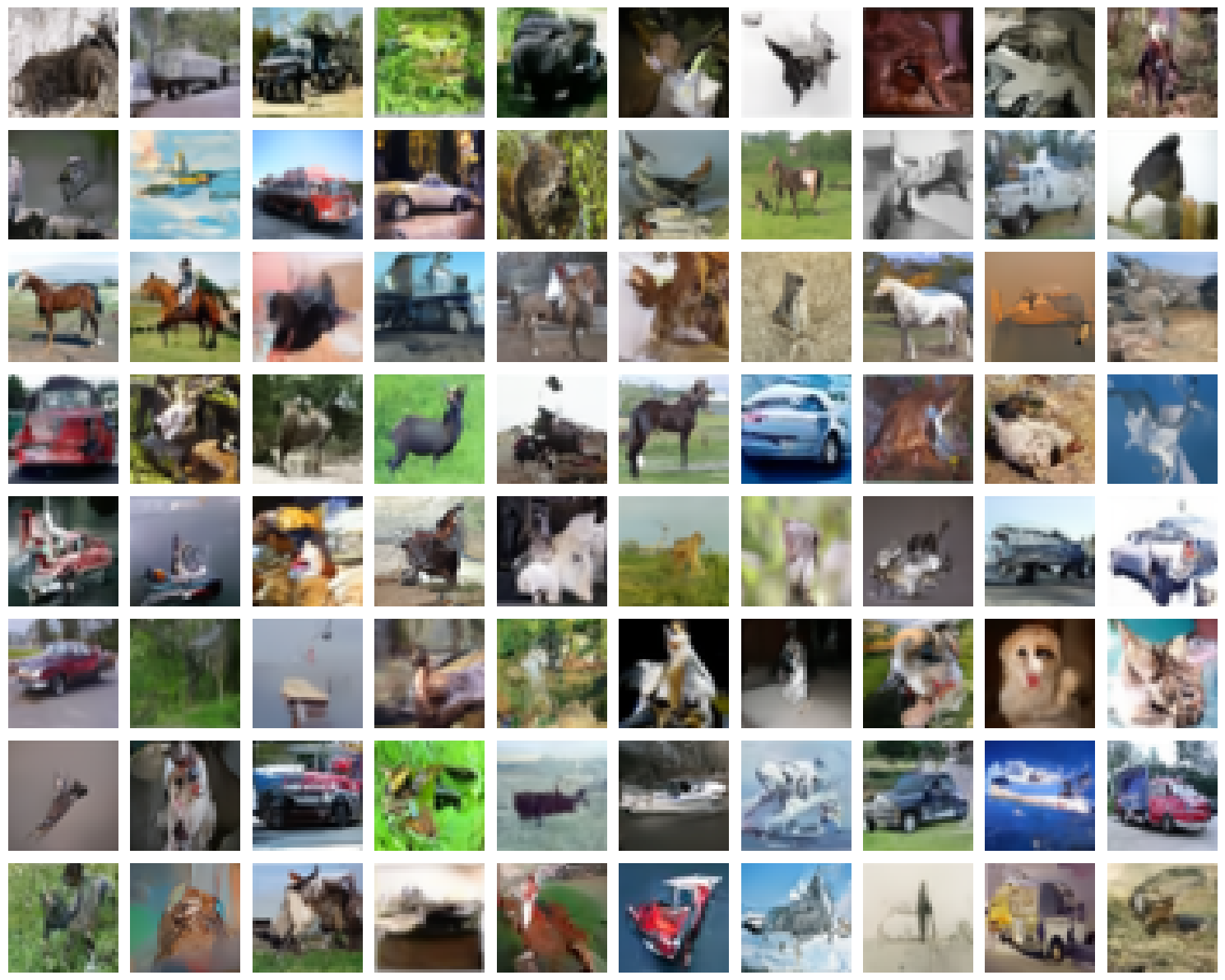}
	 \vspace{-0.8cm}
     \caption{Information scheduling + Gaussian output layer (FID=20.82).}\label{fig:7a}
   \end{minipage}\hfill
   \begin{minipage}{0.48\textwidth}
     \centering
		\includegraphics[scale=0.17]{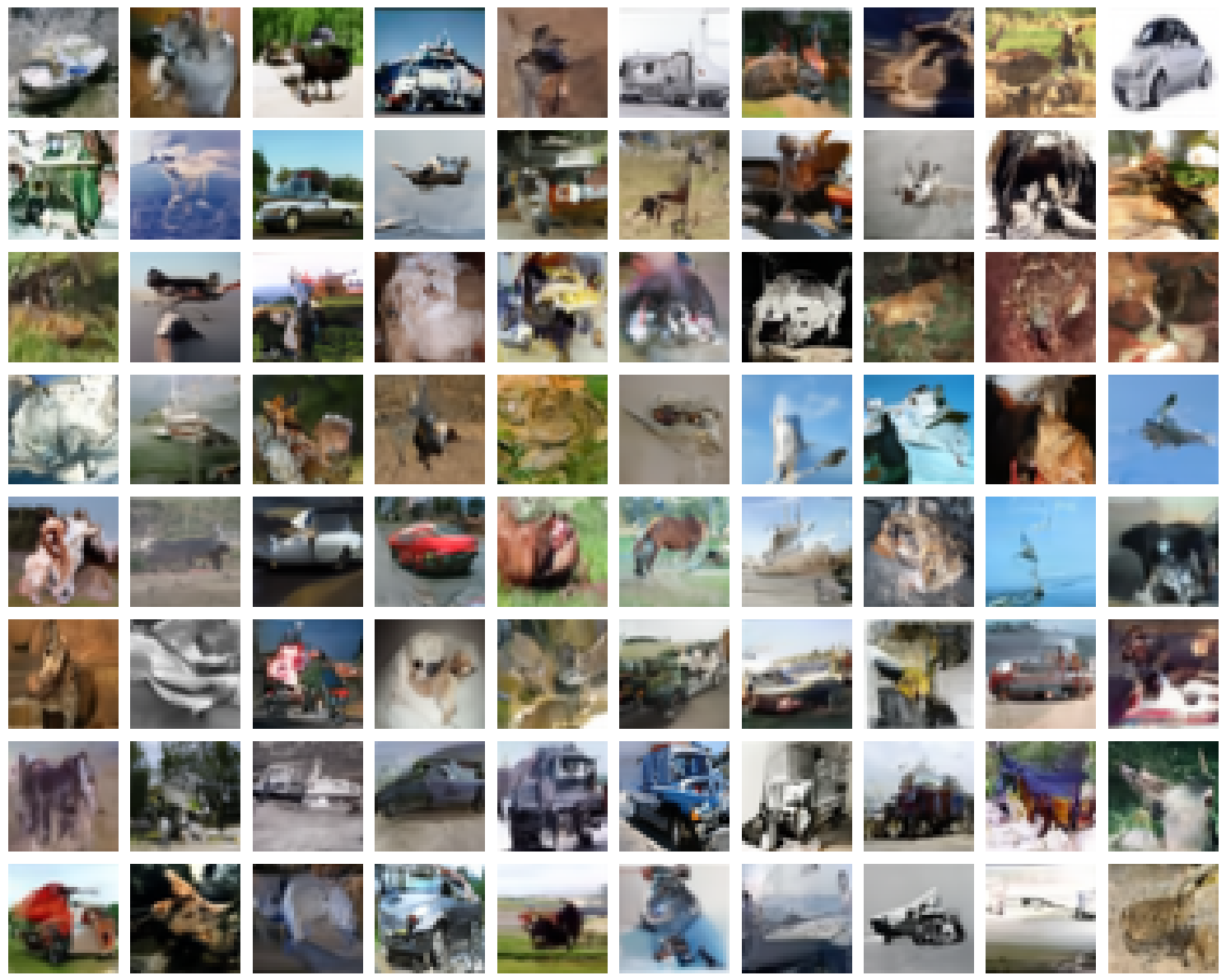}
	 \vspace{-0.8cm}
     \caption{Gaussian output layer only (FID=27.49).}\label{fig:7b}
   \end{minipage}
\medskip

   \begin{minipage}{0.48\textwidth}
     \centering
		\includegraphics[scale=0.17]{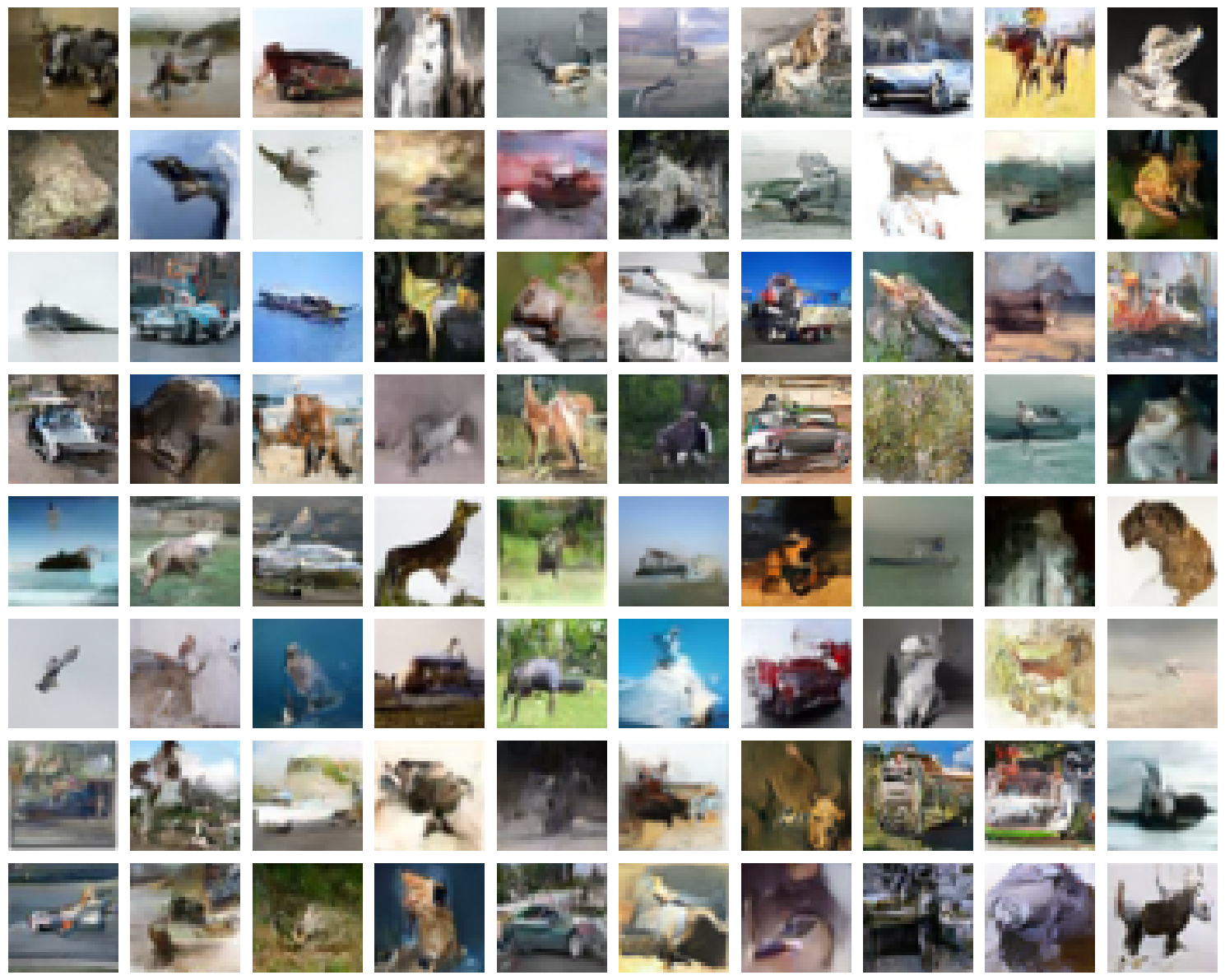}
	 \vspace{-0.8cm}
     \caption{Information scheduling only (FID=31.21).}\label{fig:7c}
   \end{minipage}\hfill
   \begin{minipage}{0.48\textwidth}
     \centering
		\includegraphics[scale=0.17]{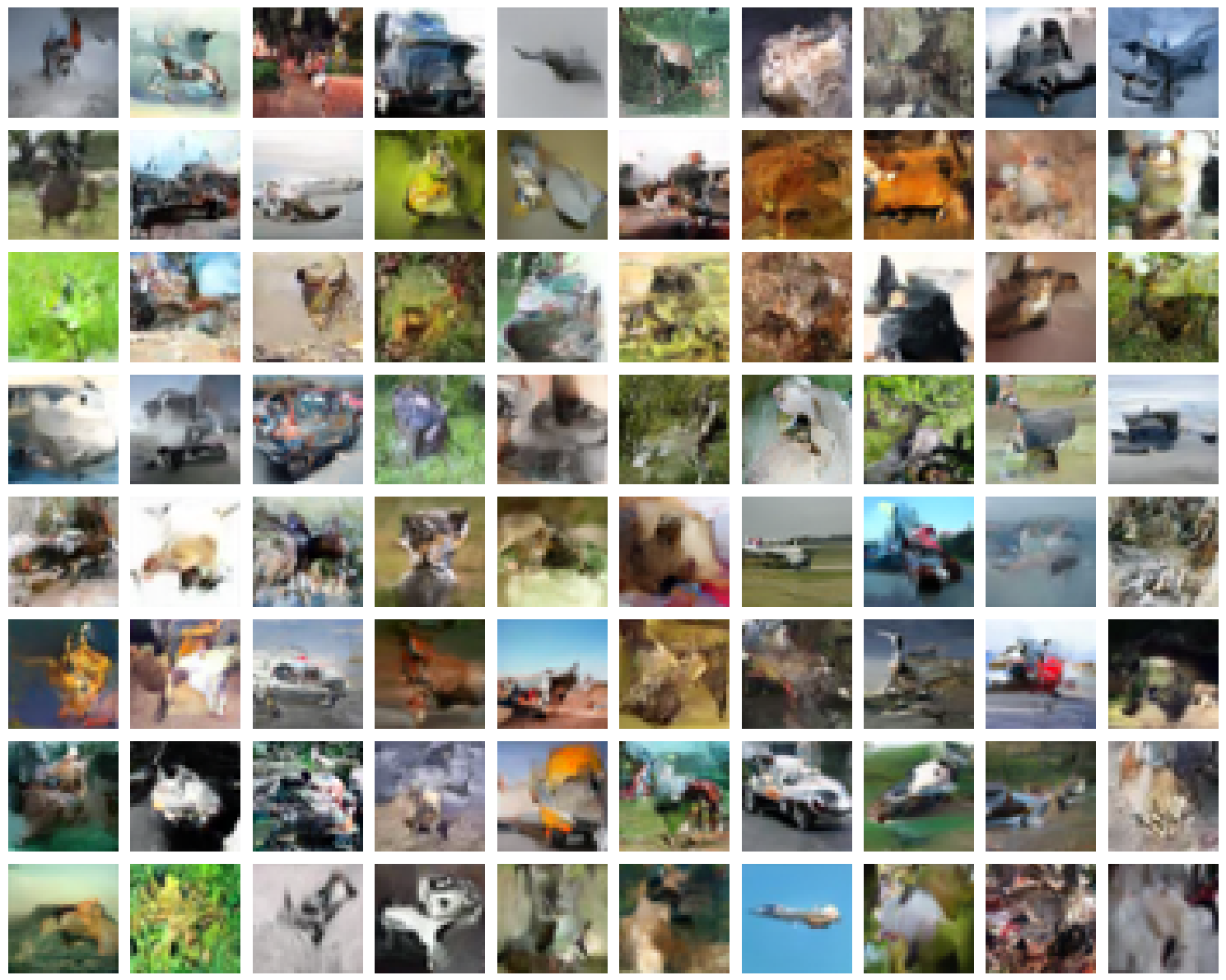}
	 \vspace{-0.8cm}
     \caption{Samples from the baseline (FID=40.66).}\label{fig:7d}
   \end{minipage}
\end{figure}
\end{subfigures}

\pagebreak

\section{ImageNet Samples}
\begin{figure}[h!]
\begin{center}
\includegraphics[scale=0.67]{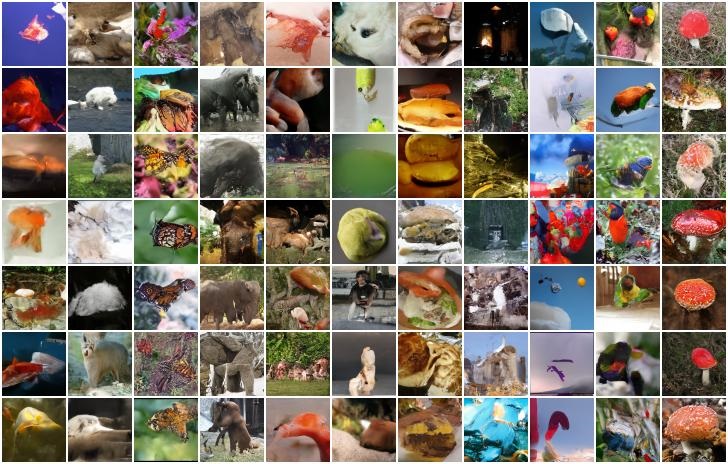}
\caption{Unguided samples from our class-conditional ImageNet $64^2$ model (FID=37.0). Classes are goldfish, arctic fox, monarch butterfly, african elephant, flamingo, tennis ball, cheeseburger, fountain, balloon, lorikeet, and agaric.}\label{fig:9}
\end{center}
\end{figure}

\begin{figure}[h!]
\begin{center}
\includegraphics[scale=0.67]{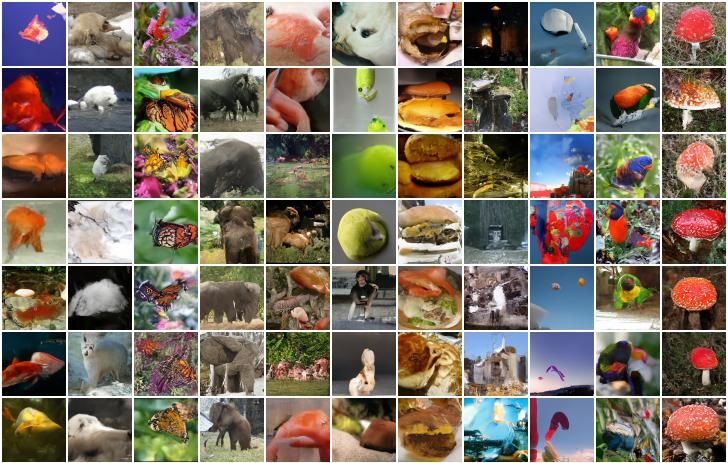}
\caption{Guided samples from our class-conditional ImageNet $64^2$ model with $\wmu=0.5, \wsigma=1.0$ (FID=27.8). Classes are goldfish, arctic fox, monarch butterfly, african elephant, flamingo, tennis ball, cheeseburger, fountain, balloon, lorikeet, and agaric.}\label{fig:10}
\end{center}
\end{figure}

\begin{figure}[h!]
\begin{center}
\includegraphics[scale=0.67]{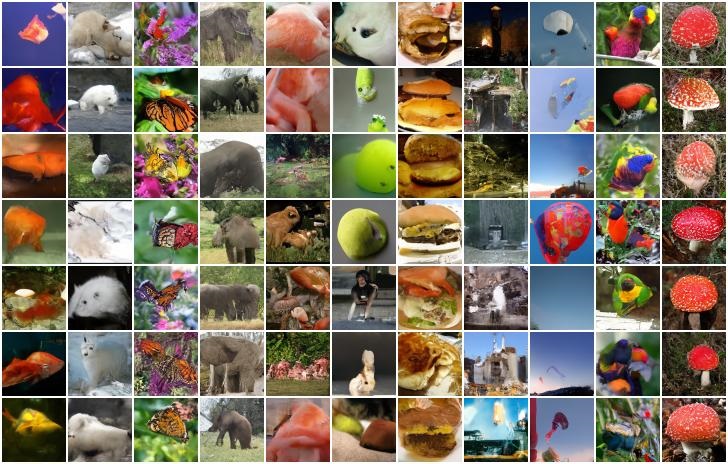}
\caption{Guided samples from our class-conditional ImageNet $64^2$ model with $\wmu=1, \wsigma=2$ (FID=21.8). Classes are goldfish, arctic fox, monarch butterfly, african elephant, flamingo, tennis ball, cheeseburger, fountain, balloon, lorikeet, and agaric.}\label{fig:11}
\end{center}
\end{figure}

\begin{figure}[h!]
\begin{center}
\includegraphics[scale=0.67]{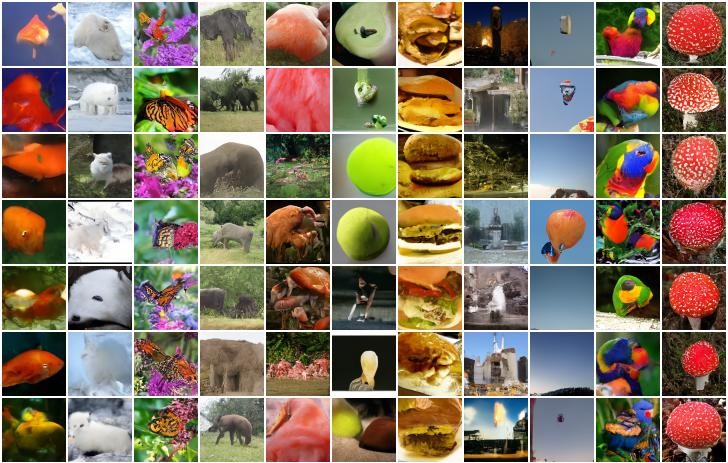}
\caption{Guided samples from our class-conditional ImageNet $64^2$ model with $\wmu=2, \wsigma=4$ (FID=17.9). Classes are goldfish, arctic fox, monarch butterfly, african elephant, flamingo, tennis ball, cheeseburger, fountain, balloon, lorikeet, and agaric.}\label{fig:12}
\end{center}
\end{figure}

\end{document}